\documentclass{article}
\usepackage{spconf,amsmath,graphicx}
\usepackage{booktabs}
\usepackage{multirow}
\usepackage{svg}

\title{Training Autoregressive Speech Recognition Models with Limited in-domain Supervision}
\name{Chak-Fai Li, Francis Keith, William Hartmann, Matthew Snover}
\address{Raytheon BBN, Cambridge MA, USA \\
{\small \tt \{chak.fai.li, francis.keith, william.hartmann, matt.snover\}@rtx.com}}
\begin{document}
\ninept
\maketitle
\begin{abstract}
Advances in self-supervised learning have significantly reduced the amount of transcribed audio required for training.
However, the majority of work in this area is focused on read speech.
We explore limited supervision in the domain of conversational speech.
While we assume the amount of in-domain data is limited, we augment the model with open source read speech data.
The XLS-R model has been shown to perform well with limited adaptation data and serves as a strong baseline.
We use untranscribed data for self-supervised learning and semi-supervised training in an autoregressive encoder-decoder model.
We demonstrate that by using the XLS-R model for pseudotranscription, a much smaller autoregressive model can outperform a finetuned XLS-R model when transcribed in-domain data is limited, reducing WER by as much as 8\% absolute.

\end{abstract}
\section{Introduction}
\label{sec:intro}

Recent advances in self-supervised learning (SSL) have led to better utilization of untranscribed data and reduced reliance on labeled data.
Some work has sought to eliminate the requirement of transcribed data entirely \cite{liu2022towards}.
Pretrained models that can be used directly or finetuned to new datasets are widely available \cite{babu2021xls, conneau2020unsupervised, baevski2020wav2vec}. 
When combined with traditional semi-supervised learning techniques, these models achieve the state-of-the-art (SOTA) on a number of datasets.

The vast majority of the work focuses on read speech, where the standard benchmark is Librispeech \cite{panayotov2015librispeech}.
While a single point of comparison has been advantageous for the community, there are other challenging applications of automatic speech recognition (ASR) beyond read speech.
It is unlikely that large scale self-supervised learning of read speech is the best way to improve conversational speech (CS) recognition; previous work has shown that pretraining in the target domain is more beneficial \cite{hsu2021robust}.

In this work we focus on reducing the amount of in-domain supervision required for autoregressive (AR) ASR models for conversational speech.
This domain presents unique challenges due to both the data and the model.
For many languages the amount of transcribed conversational speech is severely limited---even publicly available untranscribed speech is limited.
Approaches that require thousands of hours of data are difficult to apply due to lack of data.

Autoregressive models tend to require more data, at least partially due to their need to learn an internal language model (LM).
Hybrid and CTC-trained models can be finetuned on extremely small amounts of data.
While initial results may be poor, performance can be dramatically improved through the inclusion of an external lexicon and LM.
External LMs can also be applied to AR models through techniques like shallow fusion, cold fusion \cite{sriram2017cold}, deep fusion \cite{gulcehre2015using}, component fusion \cite{shan2019component}, and internal language model (ILM) estimation \cite{meng2021internal}.
However, the relative improvements from these approaches are limited compared to non-autoregressive models.

Early work with limited supervised training focused on the standard pseudolabeling approach to semi-supervised training (SST) \cite{zavaliagkos1998using, kemp1999unsupervised}.
With the advent of deep neural networks and hybrid models, there was renewed interest in SST.
The models could be trained on orders of magnitude more data \cite{parthasarathi2019lessons} and there were advances in both the selection \cite{wotherspoon2021improved} and use of the data \cite{manohar2018semi}.
Recent SST work with all-neural models include classic approaches and methods adapted from image classification.
Noisy student training \cite{park2020improved}, a commonly used technique, is an iterative pseudolabeling approach with filtering that incorporates data augmentation on the source side.
Instead of iteratively updating the pseudotranscripts, they can also be generated on-the-fly with a continuously updated transcription model \cite{higuchi2022momentum, manohar2021kaizen}.
Combining SSL and SST has further pushed the state-of-the-art \cite{zhang2022bigssl}.

Some recent work has applied these techniques to conversational speech.
In \cite{conneau2020unsupervised}, they pretrained a model on a large amount of read speech and conversational speech from the BABEL program.
When the models were finetuned on 30+ hours of in-domain data and combined with an external LM, they were able to surpass the performance of state-of-the-art hybrid models.
Later work explored finetuning pretrained models to conversational speech \cite{wiesner2022injecting}.
When the data was limited, performance was poor compared to a model trained on the full set of supervised data.
The work most similar to ours is \cite{khurana2021unsupervised}.
They start from a model trained only on read speech and attempt to adapt to a conversational dataset through SST.
However, they do not use self-supervised learning and performance on conversational speech is poor compared to a fully supervised model.

We compare the state-of-the-art for three ASR approaches where the amount of transcribed in-domain data ranges from 68 hours to nothing.
Our contributions include:

\begin{itemize}
    \item Demonstration that autoregressive models can outperform state-of-the-art hybrid models and large finetuned XLS-R models in the low-resource conversational speech domain
    \item Drastically reducing the amount of in-domain supervised data required for autoregressive conversational speech models
    \item Showing read speech can be beneficial for conversational speech systems.
\end{itemize}


\section{Limited Supervision Training for Autoregressive Models}\label{sec:method}

Autoregressive style models require large amounts of training data.
While it is possible to adapt hybrid and CTC-trained models using limited transcribed data, autoregressive models perform a more complicated task and must rely on their own internal LM.
We leverage out-of-domain data in order to bootstrap the learning process.
For out-of-domain data we use the type of transcribed data most likely to exist in any language, read speech.
If we train a model solely on the limited amounts of transcribed conversational speech we have available, the models typically fail to converge.
By adding read speech to the limited amount of conversational speech, the models converge and produce output that is competitive with a fully supervised model.

Our training pipeline is as follows.
A baseline model is trained from a random start using transcribed read speech and a limited amount of transcribed conversational telephone speech (CS).
The decoder is dropped and an intermediate layer in the encoder is used as a feature representation.
All available audio data (both transcribed and untranscribed) is encoded using the representation from the encoder.
Next, we use the HuBERT \cite{hsu2021hubert} approach for self-supervised learning (SSL) as described in \cite{li2022combining}.
The embedded features are clustered using k-means and the clusters become targets for the self-supervised learning.
During self-supervised learning only the encoder of the speech model is trained.
Since the trained model can only predict the unsupervised clusters, it cannot be used directly for ASR.
We finetune the pretrained encoder and a randomly initialized decoder using the original transcribed data.
The untranscribed data is transcribed either by this model or a preexisting model in order to produce pseudotranscripts.
The pretrained model is trained on both the trancribed data and the pseudotranscripts.
Finally, the model is further finetuned to the available in-domain transcribed data.

\section{Experimental Setup}\label{sec:setup}

\subsection{Data}\label{sec:data}

\begin{table}
    \caption{\label{tab:data} {\it Training data distribution for Kazakh and Swahili}}
    \centering
    \begin{tabular}{lcc}
    \toprule
        Data Type & Kazakh & Swahili \\
        \midrule
        Read Speech & 332 & 146 \\
        Transcribed CS & 50 & 68 \\
        Untranscribed CS & 8 & 58 \\
        Untranscribed Broadcast News & 164 & 149 \\
    \bottomrule
    \end{tabular}
\end{table}

For all of our experiments we use data provided through the IARPA MATERIAL program for the Swahili and Kazakh languages.
The terms for the datasets used during the program are provided in parentheses.
The data consists of transcribed CS data (\textit{traindev}) and approximately 200 hours of additional untranscribed data (\textit{eval}) comprised of both CS and broadcast news (BN).
Note that the transcribed CS data is identical to the data provided during the IARPA BABEL program and available through the LDC.
Exact data splits for the two languages can be seen in Table \ref{tab:data}.
In addition to the training data, the test set (\textit{analysis}) contains both CS and BN data as well.
We only report results on the CS subset as that is our focus.
We supplement the MATERIAL data with read speech for both languages.
For Swahili the data comes from CommonVoice\footnote{https://commonvoice.mozilla.org, version 8.0}\cite{ardila2019common}, while the Kazakh data is from the Kazakh Speech Corpus \cite{khassanov2020crowdsourced}, available through OpenSLR\footnote{https://www.openslr.org/102}.

For the purposes of building language models, we also use data automatically scraped from the web using the method described in \cite{zhang2015enhancing}.
While the additional LM data does not improve the conversational speech performance of a fully supervised model, it is critical for improving performance on the broadcast data.
It also drastically reduces the out-of-vocabulary (OOV) rate.

\subsection{Models}\label{sec:models}

\begin{table}
    \caption{\label{tab:models} {\it Description of the characteristics of the compared models.}}
    \centering
    \begin{tabular}{lccc}
    \toprule
        & Hybrid & AR & XLS-R \\
        \midrule
        Parameters & 25M & 43M & 965M \\
        External LM & Ngram & None & Ngram \\
        Pretraining Data & 1560 hours & 200 hours & 436k hours \\
    \bottomrule
    \end{tabular}
\end{table}

While our focus is on improving the performance of AR models with limited supervision, we also compare performance with hybrid models and finetuned XLS-R models.
Though all-neural models have surpassed hybrid models in most settings, hybrid models are still competitive on low-resource conversational speech data \cite{li2021overcoming}.
The major differences between the three models are shown in Table \ref{tab:models}.

\subsubsection{Hybrid Models}

Our hybrid systems use a Kaldi-based \cite{povey2011kaldi} DNN-HMM model trained using BBN's Sage system \cite{hsiao2016sage}.
The input features are 40-dimensional mel-filter cepstrum coefficients (MFCCs), along with a 100-dimensional ivector.
The HMM structure is the \textit{chain} structure used in \cite{povey2016purely}.
The basic structure of the DNN is identical to the TDNN-LSTM in the Switchboard setup from \cite{cheng2017exploration}.

The hybrid model is initialized from a model trained on 1560hrs of multilingual data, the same data used in \cite{keith2018optimizing}.
An ivector extractor is also trained using the same data.
The model is finetuned using the LF-MMI objective function \cite{povey2016purely} for one epoch at a constant learning rate of $3 \times 10^{-4}$, and then with sMBR \cite{vesely2013sequence} for an additional epoch at a constant learning rate of $5 \times 10^{-6}$.
Both the supervised and semi-supervised hybrid models are trained in the same manner with the same model structure; the only difference is the data used.
All of our hybrid systems use a trigram language model.

\subsubsection{Autoregressive Model}

Our AR encoder-decoder models are conformer-based \cite{gulati2020conformer} models trained in ESPNet \cite{watanabe2018espnet}.
The configuration is similar to \cite{guo2021recent}.
The encoder has 12 layers with four attention heads, an embedding dimension of 256, and a FFN dimension of 2048.
The decoder uses 6 layers with identical parameters.
In addition to the standard cross-entropy objective function, we jointly train with the CTC objective function \cite{kim2017joint}.
Our output units are characters.

During SSL, we use frame-level cross entropy to train the conformer encoder with the k-means clusters as targets.
The encoder is trained for a maximum of 80 epochs.
The minibatch size is 256 utterances, approximately 1000 seconds of audio.
We use SpecAugment \cite{park2019specaugment} during both supervised and self-supervised training.

\subsubsection{XLS-R Models}

We use the pretrained XLS-R model from \cite{babu2021xls}.
Three variants of different sizes are available, but we use the one billion parameter version in all of our experiments.
Preliminary results demonstrated the one billion parameter model outperformed the smaller 300M parameter model.
While the larger 2B parameter version gives slightly better performance, the GPU memory requirements prevented us from carrying out large scale experiments with it.

The pretrained model uses a wav2vec 2.0 \cite{baevski2020wav2vec} style contrastive loss and is trained on 436k hours of speech across 128 languages.
Note that the data used in pretraining included the BABEL data for both Swahili and Kazakh\footnote{https://catalog.ldc.upenn.edu/\{LDC2017S05,LDC2018S13\}}, which is identical to the MATERIAL training data.
We finetune the XLS-R 1B parameter model using fairseq \cite{ott2019fairseq} with a configuration similar to the ASR experiments in \cite{babu2021xls}.
We use character targets, and the model is trained with CTC \cite{graves2006connectionist}, selecting the best model based on validation set WER.
We train with a learning rate of $2 \times 10^{-5}$ using the same warm-up procedure in \cite{babu2021xls} for 100k total updates.
We used 320k max input tokens per batch, though we occasionally decreased this value to run experiments on smaller GPUs.
Empirically, we observed little difference in terms of WER when modifying the batch size, nor did we observe notable changes by modifying the total number of updates to compensate for the change in batch size.
With the exception of these changes to batch size, we use identical parameters across all XLS-R finetuning experiments for both supervised and semi-supervised finetuning.
During decoding we incorporate a lexicon and LM through a Kaldi-based WFST decoder, similar to \cite{zenkel2017comparison}.

\section{Results}\label{sec:results}

\begin{figure}[h]
\begin{minipage}[b]{1.0\linewidth}
  \centering
  \centerline{\includegraphics{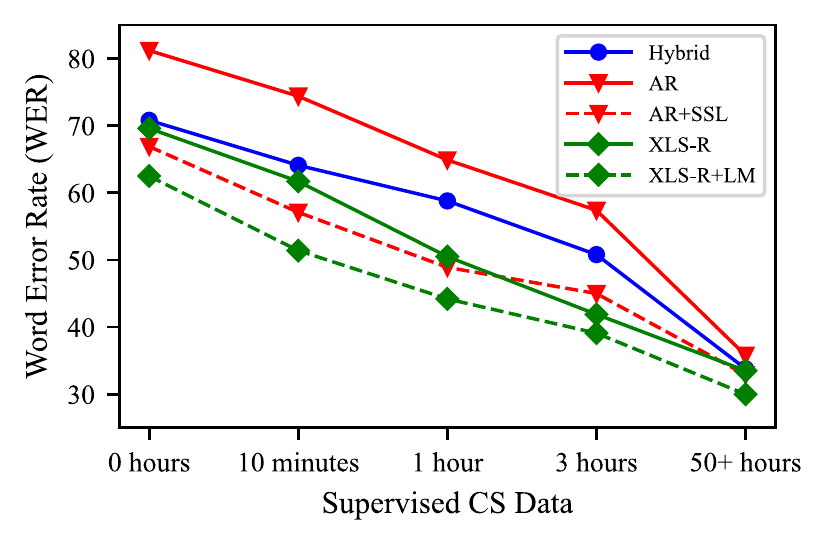}}
  \centerline{(a) Swahili}\medskip
\end{minipage}
\begin{minipage}[b]{1.0\linewidth}
  \centering
  \centerline{\includegraphics{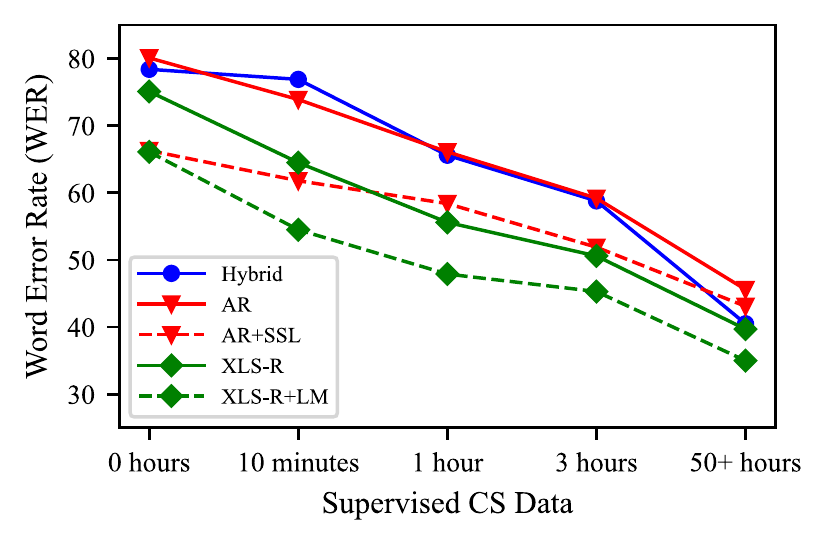}}
  \centerline{(b) Kazakh}\medskip
\end{minipage}

\caption{Results after supervised finetuning for (a) Swahili and (b) Kazakh. The hybrid and XLS-R models are always pretrained, but the AR model is shown with and without SSL. The XLS-R results include performance with and without an external LM.}
\label{fig:supervised}
\end{figure}

\subsection{Supervised Finetuning}

In Figure \ref{fig:supervised} results are presented for all three model types using various amounts of transcribed CS data for supervised finetuning.
Note that all models also include read speech in their training.
The XLS-R and hybrid models always start from a pretrained model, but we show the AR model both with and without self-supervised learning (SSL).
For the hybrid and XLS-R models the pretraining is language independent while the pretraining for the AR model uses in-domain data and is repeated for each dataset.
We first focus on the case where we have no supervised data.
Without pretraining, the AR model is practically broken; a model with a WER of 80\% is unlikely to be of use for any task.
Once the pretraining has been added, performance improves dramatically and surpasses the hybrid model.
When the amount of supervised data is limited, the use of an external LM is critical for the XLS-R model.
Without the external LM, the AR model is competitive with the XLS-R model, but the XLS-R model is clearly the best model when the external LM is added.


\begin{figure}[h]
\begin{minipage}[b]{1.0\linewidth}
  \centering
  \centerline{\includegraphics{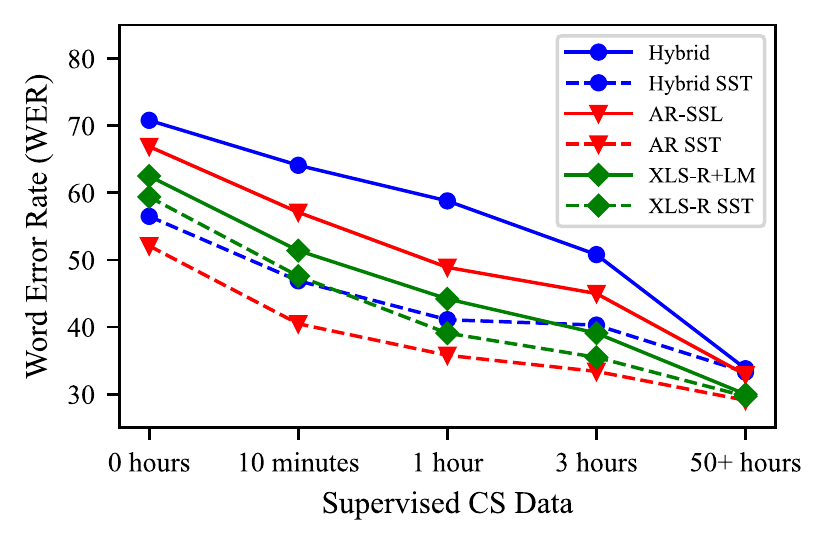}}
  \centerline{(a) Swahili}\medskip
\end{minipage}
\begin{minipage}[b]{1.0\linewidth}
  \centering
  \centerline{\includegraphics{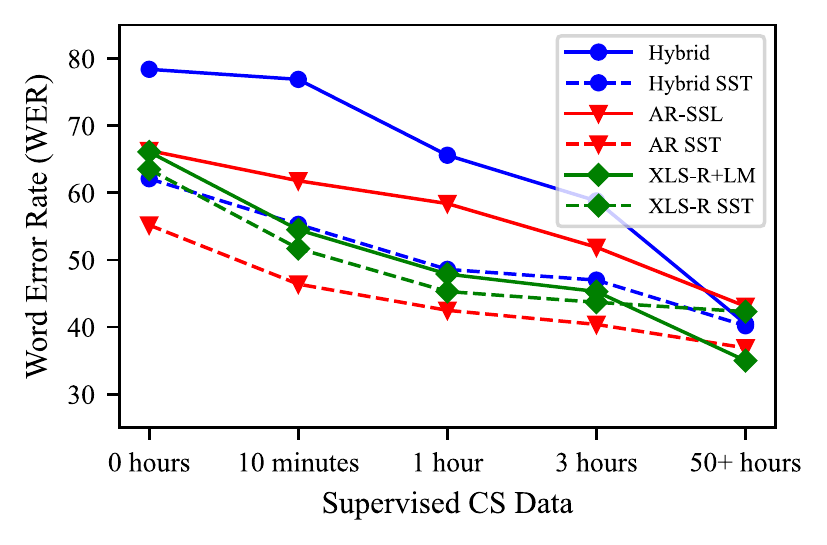}}
  \centerline{(b) Kazakh}\medskip
\end{minipage}

\caption{Results with and without semi-supervised training for (a) Swahili and (b) Kazakh. In all cases the XLS-R model with an external LM is used for generating the pseudotranscripts. Note that while the XLS-R SST result for Kazakh sees a degradation at 50 hours, there was an improvement on the broadcast subset. }
\label{fig:sst}
\end{figure}

\subsection{Semi-supervised Adaptation}

In Figure \ref{fig:sst} results are shown with and without SST.
In all cases the transcription model is an XLS-R model trained on the same amount of data as the reported model and includes an external LM.
In all cases the XLS-R model sees limited improvement from SST.
Note that the XLS-R model uses an external LM in all results in Figure \ref{fig:sst}.
Once the full amount of supervised data is presented, no model sees significant improvement from SST.
When in-domain supervision is limited, the AR model is the best model.
With 10 minutes of transcribed CS data, the AR model outperforms the XLS-R model by 7.1\% and 5.3\% absolute on Swahili and Kazakh, respectively.
With 3 hours of transcribed CS data, the AR models match the performance of hybrid models trained on 50+ hours.

\begin{table}
    \caption{\label{tab:sst} {\it Comparing the XLS-R and AR model for transcription during SST. The columns refer to the model used for generating pseudotranscriptions for SST, but the final model is always the AR model.}}
    \centering
    \begin{tabular}{lccc}
    \toprule
        Condition & None & AR & XLS-R \\
        \midrule
        Kazakh 3 hours & 51.9 & 43.3 & 40.4 \\
        Swahili 3 hours & 45.0 & 37.7 & 33.4 \\
    \bottomrule
    \end{tabular}
\end{table}

\subsection{Role of XLS-R in SST}

One additional question is whether the XLS-R model is needed when the amount of in-domain supervised data is limited.
Can we achieve similar performance by using the pretrained AR model as a transcription model during SST?
In Table \ref{tab:sst} we directly compare the two transcription models.
Note that when we use the AR model for transcription, we do not use the internal decoder.
Instead we only use the CTC output of the encoder and then compose the output with a lexicon and LM as in \cite{li2022combining}.
While the AR model still sees significant improvement when using itself for transcription, there is a significant gain from switching to the XLS-R model.
With limited amounts of supervision the AR model is the best overall model, but it still benefits from the XLS-R model for pseudotranscription.

\subsection{Relevance of Read Speech}

\begin{table}
    \caption{\label{tab:read} {\it Comparing WER performance with and without read speech for the XLS-R model on Swahili. Results do not include an external LM during decoding and we do not consider SST.}}
    \centering
    \begin{tabular}{lcc}
    \toprule
        Supervised CS Data & CS Only & CS + Read Speech \\
        \midrule
        0 hours & --- & 69.6 \\
        10 minutes & \textgreater 100 & 61.7 \\
        1 hour & 60.3 & 50.5 \\
        3 hours & 45.8 & 41.9 \\
        68 hours & 30.9 & 33.5 \\
    \bottomrule
    \end{tabular}
\end{table}

The use of read speech is critical to our results.
If we remove the read speech, then AR models with limited amounts of in-domain supervision will fail to train at all.
Obviously the condition without any conversational speech is no longer feasible as there is no supervision.
With the XLS-R model it is at least feasible to train the model on the limited amounts of supervised CS data only.
We report results on Swahili in Table \ref{tab:read}.
When limiting the amount of CS data to 10 minutes, the WER of the final model is over 100\%.
Adding the read speech to the training process brings the WER to 61.7\%.
It is clear that when the two types of data are combined, both contribute to the improved performance of the model.
At one hour of supervised CS data, the model is able to function without read speech, but still improves by about ten points absolute when it is added.
Only when the model has access to sufficient amounts of CS data does the read speech no longer provide a benefit.
With all of the supervised CS data, read speech actually hurts performance.
Note a strong LM reduces, but does not eliminate the need for read speech.

\subsection{Speakers in Transcribed Data}

\begin{table}
    \caption{\label{tab:speaker} {\it Comparing the effect of the number of speakers on WER in the 3 hour supervision set using the AR model on Kazakh.}}
    \centering
    \begin{tabular}{lcc}
    \toprule
        Training Stage & 38 Speakers & 300 Speakers \\
        \midrule
        Supervised Only & 60.4 & 59.2 \\
        + SSL & 52.4 & 51.9 \\
        + SST & 40.9 & 40.4 \\
    \bottomrule
    \end{tabular}
\end{table}

When creating the limited supervision training sets, we randomly selected utterances from the full training set.
For the 3 hour set, this resulted in 300 speakers for training.
This is unrealistic with respect to real world scenarios.
A transcriber is more likely to transcribe longer amounts from fewer speakers; transcribing at the utterance level is more challenging and time-consuming.
As an ablation, we selected a second 3 hour set by speaker.
While it contained the same amount of transcribed audio, it only had 38 unique speakers.
Results in Table \ref{tab:speaker} show the number of speakers have a negligible effect.

\section{Conclusions}\label{sec:conclusion}

We have demonstrated that we can train an AR model on limited amounts of in-domain transcribed CS data by leveraging out-of-domain read speech and SSL on additional in-domain data.
Overall our model is less than 5\% the size of a finetuned XLS-R model and trained on less than 1\% of the data.
Our results demonstrate that a smaller model targeted to a particular domain can outperform a larger model trained on a vast amount of diverse data.
However, as the amount of supervised in-domain data grows, the performance of the XLS-R model begins to match our AR model.
In previous work \cite{li2022combining} we were only able to match the performance of hybrid models with our AR models.
Not only do we now surpass hybrid models by a large margin, but we demonstrate that an AR model trained on only three hours of in-domain supervised data can match the performance of a hybrid model trained with 50+ hours.

Our experiments used data and resources that are only partially publicly available.
The webdata was collected using a proprietary tool and only a subset of the MATERIAL data is available (as part of BABEL data releases).
We plan to develop an experimental setup using only publicly accessible data in the future so that other researchers can more easily work in this important application area.

\section{ACKNOWLEDGEMENTS}


This work was supported by the Intelligence Advanced Research Projects Activity (IARPA) via Department of Defense US Air Force Research Laboratory contract number FA8650-17-C-9118.
This document does not contain technology or Technical Data controlled under either the U.S. International Traffic in Arms Regulations or the U.S. Export Administration Regulations.

\vfill\pagebreak

\bibliographystyle{IEEEbib} 
{\footnotesize
\bibliography{refs}}

\end{document}